\documentclass[lettersize,journal]{IEEEtran}
\usepackage{amsmath,amsfonts}
\usepackage{algorithmic}
\usepackage{array}
\usepackage[caption=false,font=normalsize,labelfont=sf,textfont=sf]{subfig}
\usepackage{textcomp}
\usepackage{stfloats}
\usepackage{url}
\usepackage{verbatim}
\usepackage{graphicx}
\usepackage{color}
\usepackage{multirow}
\usepackage{float}
\usepackage{hyperref}
\usepackage{bm}
\usepackage{amsmath}
\usepackage[linesnumbered,ruled]{algorithm2e}
\usepackage{amsthm,amsmath,amssymb}\usepackage{mathrsfs}
\usepackage{epstopdf}
\usepackage[table]{xcolor}
\usepackage{amsmath,amssymb,amsthm}
\usepackage{adjustbox}
\usepackage{booktabs}   
\usepackage{multirow}   
\usepackage{makecell}   
\usepackage{graphicx}   

\makeatletter
\renewcommand{\maketag@@@}[1]{\hbox{\m@th\normalsize\normalfont#1}}%
\makeatother
\def\BibTeX{{\rm B\kern-.05em{\sc i\kern-.025em b}\kern-.08em
    T\kern-.1667em\lower.7ex\hbox{E}\kern-.125emX}}
\usepackage{balance}

\begin{document}
\title{
LLM-Guided Diagnostic Evidence Alignment for Medical Vision–Language Pretraining under Limited Pairing
}
\author{Huimin Yan, Liang Bai, Xian Yang, Long Chen

\IEEEcompsocitemizethanks{\IEEEcompsocthanksitem
Huimin Yan and Liang Bai are with Institute of Intelligent Information Processing, Shanxi University, Taiyuan, 030006, China. (Corresponding author: Liang Bai). 
E-mail: yanhm0925@163.com, bailiang@sxu.edu.cn

Xian Yang is  with Alliance Manchester Business School, The University of Manchester, Manchester, M13 9PL, UK. 
E-mail: xian.yang@manchester.ac.uk

Long Chen is with Department of Computer Science and Engineering, The Hong Kong University of Science and Technology, Hong Kong, China.
E-mail: longchen@ust.hk

}
\thanks{}
}

\markboth{}%
{How to Use the IEEEtran \LaTeX \ Templates}

\maketitle

\begin{abstract}
Most existing CLIP-style medical vision--language pretraining methods rely on global or local alignment with substantial paired data. However, global alignment is easily dominated by non-diagnostic information, while local alignment fails to integrate key diagnostic evidence. As a result, learning reliable diagnostic representations becomes difficult, which limits their applicability in medical scenarios with limited paired data. To address this issue, we propose an LLM-Guided Diagnostic Evidence Alignment method (LGDEA), which shifts the pretraining objective toward evidence-level alignment that is more consistent with the medical diagnostic process. Specifically, we leverage LLMs to extract key diagnostic evidence from radiology reports and construct a shared diagnostic evidence space, enabling evidence-aware cross-modal alignment and allowing LGDEA to effectively exploit abundant unpaired medical images and reports, thereby substantially alleviating the reliance on paired data. Extensive experimental results demonstrate that our method achieves consistent and significant improvements on phrase grounding, image--text retrieval, and zero-shot classification, and even rivals pretraining methods that rely on substantial paired data.

\end{abstract}


\section{Introduction}
\IEEEPARstart{M}{edical} vision--language pretraining (VLP) aims to learn transferable multimodal representations by modeling the correspondences between medical images and clinical reports \cite{DBLP:conf/icml/Zhu00ZWY25}, which plays an important role in enabling clinically meaningful diagnostic decision-making, and is therefore widely applied to downstream tasks such as disease classification and image--report retrieval \cite{DBLP:conf/cvpr/ZouY25a, DBLP:journals/tnn/LuLPDH25}. Early medical VLP methods typically adapt the CLIP framework to the medical domain \cite{DBLP:conf/cvpr/ZhangYCYY25}, employing global contrastive objectives to align entire medical images with their corresponding full clinical reports on substantial paired data, thereby learning transferable multimodal representations \cite{DBLP:conf/icml/GijsenR25, li2025enhancing}. This paradigm implicitly assumes that global alignment is sufficient to capture diagnostic semantics. 
\begin{figure}[t]
	\centering
	\includegraphics[scale=.58]{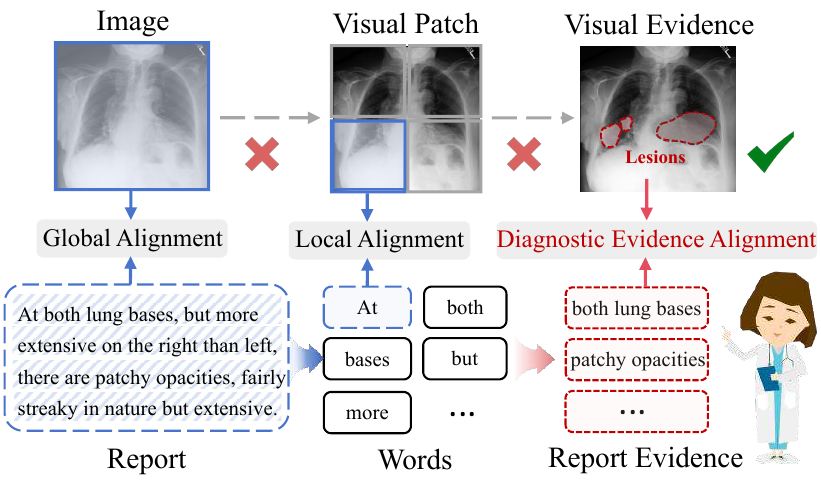}
	\caption{
Motivation of LGDEA.
Global and local alignment may overlook diagnostic evidence,
whereas LGDEA aligns images and reports in a shared diagnostic evidence space.
}
	\label{fig:motivation}
\end{figure}

However, radiology reports often describe multiple localized pathological findings, while diagnostically critical visual lesions are typically subtle and spatially localized. To mitigate this issue, some methods introduce local alignment by matching report words or phrases with image regions. Such local alignment methods partially alleviate the limitation of global alignment in modeling localized diagnostic cues, enabling the model to attend to image regions that are relevant to pathological descriptions \cite{MedKLIP, li2024mlip, MedAligner}.

So far, state-of-the-art global and local alignment methods share two common characteristics: 1) They heavily rely on substantial and high-quality paired data to provide sufficient cross-modal supervisory signals. 2) They essentially follow a feature-level alignment paradigm, which is well suited for scenarios such as natural image–text pairs where semantic correspondences are relatively explicit. However, in the medical domain, high-quality paired data are inherently limited, and diagnostic semantics are often implicit in the integrated reasoning over a small number of diagnostic evidence, which cannot be adequately captured by feature similarity alone ~\cite{xie2024unimiss+, zhang2025bridging}. Consequently, as illustrated in Figure~\ref{fig:motivation}, under limited paired supervision in medical settings, global alignment methods tend to overlook fine-grained diagnostic details \cite{DBLP:conf/icml/Yang00W24}, allowing non-diagnostic information to dominate the learned embeddings, whereas local alignment methods overemphasize local correspondences, shifting the learning objective toward low-level visual details and making it difficult to acquire high-level diagnostic evidence that underpin clinical decision-making \cite{AFLoc}.

Based on these observations, we argue that effective medical VLP under limited pairing should follow the real diagnostic process of clinicians, shifting from feature-level alignment to \emph{diagnostic evidence alignment}.
However, performing diagnostic evidence alignment under limited paired data poses \textbf{two key challenges}:
(1) How to construct a reliable diagnostic evidence space. (2) How to effectively learn diagnostic evidence alignment when paired data are scarce, but additional unimodal data are available without explicit cross-modal correspondences.
To address these challenges, we propose \textbf{LGDEA}, an LLM-Guided Diagnostic Evidence Alignment pretraining framework that leverages medical knowledge to compensate for limited paired data. 
Specifically, we first utilize large language models (LLMs) to extract key diagnostic evidence from radiology reports and organize them into a structured cross-modal diagnostic evidence space.
Then, report-derived evidence from limited paired data are used to guide lesion-level visual representation learning, aligning visual evidence within the diagnostic evidence space.
Finally, based on evidence-centric representations, we infer cross-modal relations for unimodal data by propagating sparse paired supervision over image--image and text--text evidence graphs, thereby guiding evidence-aware cross-modal alignment over both limited paired data and abundant unpaired images and reports.
Our main contributions are summarized as follows:
\begin{itemize}
\item 
We leverage LLMs to construct a cross-modal diagnostic evidence space, establishing a diagnostic evidence foundation for medical multimodal pretraining.

\item 
We propose an evidence-aware cross-modal alignment method that effectively leverages abundant unpaired medical images and reports, thereby substantially reducing the reliance on paired data.

\item 
Extensive experiments show that our method consistently improves performance on phrase grounding, image--text retrieval, and zero-shot classification, even outperforming methods trained with more paired data.
\end{itemize}

\section{Related Work}
\subsection{Global Alignment in Medical VLP}
Inspired by the CLIP-style contrastive learning framework \cite{CLIP, chen2024medblip}, numerous studies aligns medical images and their corresponding reports by projecting them into a shared embedding space. PubMedCLIP \cite{eslami2023pubmedclip} adapts CLIP to medical data to improve medical visual question answering, while MedCLIP \cite{MedCLIP} performs alignment at the disease level using similarity matrices. CARZero \cite{CARZero} further enhances alignment accuracy by introducing cross-attention between image and report features to capture more complex semantic relationships. 
Despite these advances, existing medical vision–language pre-training methods largely rely on global representation alignment, overlooking the localized and multi-lesion nature of radiological data \cite{lian2025efficient}. Under limited paired supervision, such strategies often damage diagnostic signals, allowing non-diagnostic information to dominate the embedding space and weakening clinically meaningful semantics.

\subsection{Local Alignment in Medical VLP}
To address the limitations of global contrastive learning in capturing localized pathology, prior work has explored fine-grained vision–language alignment strategies \cite{BioViL, liang2025medfilip, yu2025fine}. GLoRIA \cite{huang2021gloria} and MGCA \cite{MGCA} incorporate local or multi-granularity contrastive objectives to align words or tokens with attention-weighted image regions, while MedKLIP \cite{MedKLIP} and MedAligner  \cite{MedAligner} leverage structured reports or explicit word–region alignment to enhance local semantic matching. AFLoc \cite{AFLoc} further introduces multi-level semantic contrastive learning for unlabeled pathological localization.
However, these approaches mainly focus on region-level correspondences \cite{chen2025large}, whereas medical diagnosis relies on identifying and integrating key diagnostic evidence at the sample level. Under limited paired supervision, neither global alignment nor isolated local matching is sufficient, motivating a shift toward diagnostic evidence–level alignment.

\section{Method}
\subsection{Problem Settings}
\label{sec:problem_setting}

We consider a realistic medical vision--language pretraining setting where only a small subset of images and radiology reports are paired, while the majority are unpaired.
Let $\mathcal{D}_p = \{(\mathbf{I}_p, \mathbf{R}_p)\}_{p=1}^{N_p}$ denote paired samples, and $\mathcal{D}_u^{I}= \{\mathbf{I}_u\}_{u=1}^{N_u^{I}}$ and $\mathcal{D}_u^{R}= \{\mathbf{R}_u\}_{u=1}^{N_u^{R}}$ denote unpaired images and reports.
We extract representations using an image encoder $f_{\text{image}}$ ~\cite{DosovitskiyB0WZ21} and a text encoder $f_{\text{text}}$ ~\cite{BioClinicalBERT}: 
$f_{\text{image}}: \mathbf{I} \rightarrow (\mathbf{I}^g, \mathbf{I}^l), \quad f_{\text{text}}: \mathbf{R} \rightarrow (\mathbf{R}^g, \mathbf{R}^l)$, 
where $\mathbf{I}^g$ and $\mathbf{R}^g$ denote global embeddings, and $\mathbf{I}^l$ and $\mathbf{R}^l$ denote local patch- and token-level embeddings.

\begin{figure*}[!htbp]
	\centering
		\includegraphics[scale=.5]{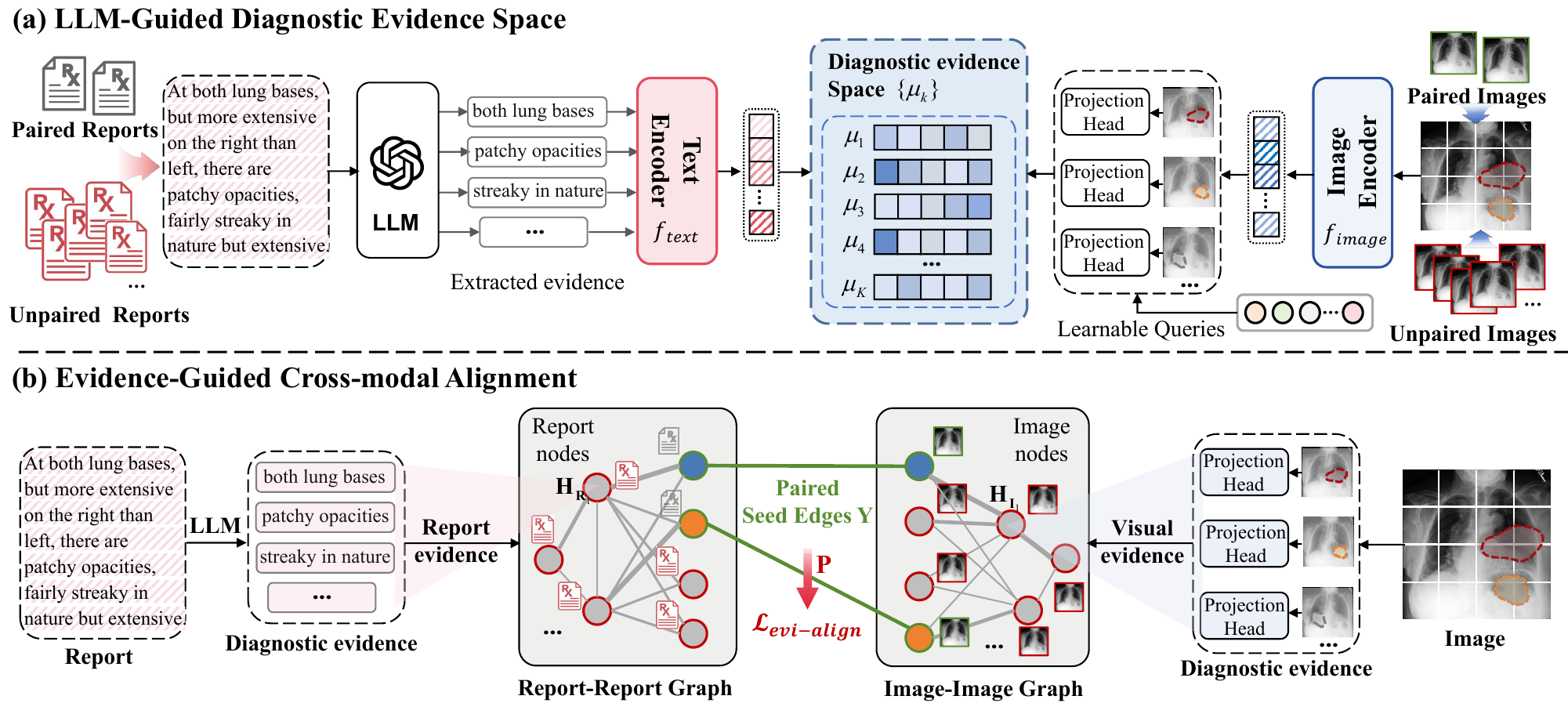}
	\caption{
    Overview of the proposed LGDEA framework.
    (a) LLMs extract diagnostic evidence from radiology reports, and both report evidence and lesion-level visual cues are projected into a shared diagnostic evidence space.
    (b) Under limited pairing, paired evidence links are used as seed edges to align report and image nodes, while report--report and image--image graphs propagate evidence-aware relations to leverage abundant unpaired reports and images for vision--language pretraining.}
	\label{fig:overview}
\end{figure*}

Existing CLIP-style pretraining methods rely on global contrastive alignment over substantial paired data, which is formulated as:
\begin{equation}
\mathcal{L}_g^{R \leftarrow I} = - \frac{1}{B} \sum_{p=1}^B \log \frac{\exp (\mathbf{I}_p^g \cdot \mathbf{R}_p^g / \tau_1)}{\sum_{k=1}^B \exp (\mathbf{I}_p^g \cdot \mathbf{R}_k^g / \tau_1)},
\label{eq1}
\end{equation}
where $B$ is the batch size and $\tau_1$ is a temperature parameter. The dot operator $\cdot$ refers to the inner product of vectors. Similarly, the loss for aligning text to images, $\mathcal{L}_g^{I \leftarrow R}$, is defined symmetrically to $\mathcal{L}_g^{R \leftarrow I}$. 
The total global alignment loss is the sum of these two components
$\mathcal{L}_g = \mathcal{L}_g^{R \leftarrow I} + \mathcal{L}_g^{I \leftarrow R}$.

Under limited paired data, global alignment often fails to capture sparse and localized diagnostic information, while local alignment tends to overemphasize fragmented details. Therefore, effective medical vision–language pretraining should follow the actual clinical diagnostic process, shifting from feature-level alignment to diagnostic evidence alignment. However, evidence-level alignment under scarce paired supervision is non-trivial, as it requires building a reliable diagnostic evidence space and learning effective cross-modal alignment by leveraging abundant unpaired unimodal data as shown in Figure~\ref{fig:overview}.

\subsection{LLM-guided Diagnostic Evidence Space}
\label{sec:text}

We leverage LLMs to construct a structured cross-modal diagnostic evidence space from data with limited pairing and abundant unpaired samples, providing a diagnostic evidence foundation for medical multimodal pretraining. First, we extract diagnostic evidence from reports and organize them into a modality-shared diagnostic evidence space:
\begin{equation}
\mathcal{E}(\mathbf{R}) = \{ e_n \}_{n=1}^{N_R} = \mathrm{LLM}(\mathbf{R}),
\end{equation}
where each evidence phrase $e_n$ corresponds to a concise and clinically meaningful abnormal finding. 
Given an evidence phrase $e_n$, we encode it into a continuous embedding using a text encoder $\mathbf{z}_n = f_{\text{text}}(e_n)$.
Second, we introduce a set of learnable diagnostic prototypes:
\begin{equation}
\mathcal{C} = \{ \boldsymbol{\mu}_k \}_{k=1}^{K}, \qquad \boldsymbol{\mu}_k \in \mathbb{R}^{d}.
\end{equation}
Each prototype represents a latent diagnostic concept that may be shared across different reports and images.
We compute a soft assignment over the diagnostic prototypes:
\begin{equation}
p(k \mid \mathbf{z}_n)
=
\frac{\exp(\mathbf{z}_n^\top \boldsymbol{\mu}_k / \tau_t)}
{\sum_{j=1}^{K} \exp(\mathbf{z}_n^\top \boldsymbol{\mu}_j / \tau_t)},
\end{equation}
where $\tau_t$ is a temperature parameter.
We jointly learn the text encoder and diagnostic prototypes by encouraging each evidence embedding to be reconstructed from the prototype space:
\begin{equation}
\mathcal{L}_{\text{rec}}
=
\sum_{e_n \in \mathcal{E}(R)}
\left\|
\mathbf{z}_n -
\sum_{k=1}^{K} p(k \mid \mathbf{z}_n)\boldsymbol{\mu}_k
\right\|_2^2
+
\sum_{k=1}^{K} \|\boldsymbol{\mu}_k\|_2^2.
\end{equation}
Through this mechanism, unpaired reports can effectively participate in representation learning by providing weak semantic supervision at the evidence level. This diagnostic evidence space provides a foundation for cross-modal learning of visual evidence.

Medical images often contain multiple localized pathological regions whose visual manifestations are subtle, spatially sparse, and difficult to annotate. Under limited pairing, global alignment is insufficient to guide models to focus on such diagnostically critical regions.

We therefore ground visual learning in the shared diagnostic evidence space established from text, enabling evidence-level alignment between images and reports.
First, we introduce a set of $L$ learnable lesion queries $\{\mathbf{q}_\ell\}_{\ell=1}^{L}$ that attend to patch-level visual features.
Given a medical image $\mathbf{I}$, a vision encoder produces patch embeddings $\mathbf{I}^l \in \mathbb{R}^{P \times d_v}$.
Each lesion query interacts with the patch features through a query-based attention mechanism:
$\mathbf{v}_{\ell}
=
\mathrm{Attn}(\mathbf{q}_\ell, \mathbf{I}^l, \mathbf{I}^l),
\quad \ell = 1, \dots, L,$
yielding a set of lesion-level embeddings $\mathbf{v}_{\ell} \in \mathbb{R}^{d_v}$.
These embeddings represent latent visual evidence corresponding to potential pathological regions.

To enable evidence-level alignment with text, each lesion embedding is projected into the shared diagnostic prototype space learned from reports:
\begin{equation}
\mathbf{Q}_I(\ell, k)
=
\frac{\exp(\phi(\mathbf{v}_{\ell})^\top \boldsymbol{\mu}_k / \tau_p)}
{\sum_{j=1}^{K} \exp(\phi(\mathbf{v}_{\ell})^\top \boldsymbol{\mu}_j / \tau_p)},
\end{equation}
where $\boldsymbol{\mu}_k$ denotes the $k$-th diagnostic prototype.
The resulting distribution $\mathbf{Q}_I(\ell,\cdot)$ characterizes the diagnostic evidence associated with each lesion. Lesion-level prototype distributions are aggregated to obtain an image-level diagnostic distribution:
\begin{equation}
\bar{\mathbf{Q}}_I
=
\frac{1}{L}
\sum_{\ell=1}^{L} \mathbf{Q}_I(\ell, \cdot).
\end{equation}

For images paired with radiology reports, report-derived diagnostic evidence provide explicit semantic guidance.
Given a report $\mathbf{R_p}$ paired with an image $\mathbf{I_p}$, we aggregate prototype assignments of all extracted report evidence to form a report-induced diagnostic distribution:
\begin{equation}
\bar{\mathbf{Q}}_R(k)
=
\frac{1}{|\mathcal{E}(\mathbf{R_p})|}
\sum_{e_n \in \mathcal{E}(\mathbf{R_p})} p(k \mid \mathbf{z}_n).
\end{equation}

We then align visual evidence with textual evidence by minimizing the KL divergence:
\begin{equation}
\mathcal{L}_{p}^{evid}
=
\mathrm{KL}\!\left(
\bar{\mathbf{Q}}_R
\,\|\, 
\bar{\mathbf{Q}}_I
\right),
\end{equation}
where the report-induced distribution $\bar{\mathbf{Q}}_R$ serves as a teacher signal.
This evidence-level distillation enables paired images to learn lesion-level visual evidence consistent with diagnostic semantics, without requiring explicit lesion annotations.

For unpaired images, report-derived supervision is unavailable.
However, unpaired images still exhibit meaningful visual structure and similarity patterns, which can be exploited to stabilize evidence learning.
Specifically, we collect all lesion instances from both paired and unpaired images within a mini-batch as
$\{\mathbf{v}_i, \mathbf{Q}_i\}_{i=1}^{N_L}$, where $N_L = B \times L$.
We measure visual similarity using cosine similarity between lesion embeddings and define the $k$-nearest neighbors as
$\mathcal{N}_k(i) = \operatorname{TopK}_{j \neq i}\big(\cos(\mathbf{v}_i, \mathbf{v}_j)\big)$.
To encourage evidence-level consistency, we align the diagnostic prototype distributions of visually similar lesions using a similarity-weighted KL divergence:
\begin{equation}
\mathcal{L}_{u}^{evid}
=
\frac{1}{N_L}
\sum_{i=1}^{N_L}
\sum_{j \in \mathcal{N}_k(i)}
w_{ij}\,
\mathrm{KL}\!\left(
\mathbf{Q}_i
\,\|\, 
\mathrm{stopgrad}(\mathbf{Q}_j)
\right),
\end{equation}
where $w_{ij}$ is computed by a softmax over cosine similarities.
This consistency constraint allows unpaired images to actively participate in training by absorbing diagnostic semantics from nearby, evidence-aligned lesions, thereby anchoring their lesion-level representations within the shared diagnostic evidence space.

\subsection{Evidence-guided Cross-modal Alignment}
\label{sec:relation}

To enable medical vision--language pretraining under limited paired data,
we propose an evidence-guided weakly-supervised alignment strategy.
The key challenge is that when explicit paired relations are scarce or even absent,
direct image--report alignment becomes unreliable.
Instead, we aim to learn higher-order cross-modal relations that reflect
evidence-level semantic relatedness beyond direct pairing, and use them as weak
supervision for training.

Given a mini-batch of $B$ images and $B$ reports, we define an evidence-aware
cross-modal alignment loss as
\begin{equation}
\mathcal{L}_{\text{evi-align}}
=
\mathcal{L}_{\text{evi}}^{R \leftarrow I}
+
\mathcal{L}_{\text{evi}}^{I \leftarrow R},
\end{equation}
where the image-to-report direction is
\begin{equation}
\mathcal{L}_{\text{evi}}^{R \leftarrow I}
=
-\frac{1}{B}
\sum_{i=1}^{B}
\sum_{j=1}^{B}
\mathbf{P}_{ij}
\log
\frac{
\exp \left( \mathbf{H}_{\mathbf{I}_i}^\top \mathbf{H}_{\mathbf{R}_j} / \tau_2 \right)
}{
\sum_{k=1}^{B}
\exp \left( \mathbf{H}_{\mathbf{I}_i}^\top \mathbf{H}_{\mathbf{R}_k} / \tau_2 \right)
},
\label{eq:evi_align_ir}
\end{equation}
and $\mathcal{L}_{\text{evi}}^{I \leftarrow R}$ is defined symmetrically.
Here, $\tau_2$ is the temperature hyperparameter, and
$\mathbf{P}_{ij} \in [0,1]$ denotes the higher-order evidence relation
between image $\mathbf{I}_i$ and report $\mathbf{R}_j$.
Unlike hard paired labels, $\mathbf{P}_{ij}$ provides soft supervision by
measuring how likely two samples share similar diagnostic evidence.

For each report $\mathbf{R}$, we extract a set of diagnostic evidence
$\mathcal{E}(\mathbf{R})$ using an LLM, and obtain its report-level diagnostic
representation by aggregating evidence embeddings:
\begin{equation}
\mathbf{H}_{\mathbf{R}}
=
\frac{1}{|\mathcal{E}(\mathbf{R})|}
\sum_{e_n \in \mathcal{E}(\mathbf{R})}
\mathbf{z}_n
\in \mathbb{R}^{d}.
\end{equation}
Similarly, for an image $\mathbf{I}$ with $L$ lesion queries, we aggregate lesion-level visual evidence projected into the diagnostic evidence embedding space:

\begin{equation}
\mathbf{H}_{\mathbf{I}}
=
\frac{1}{L}
\sum_{\ell=1}^{L}
\phi(\mathbf{v}_\ell)
\in \mathbb{R}^{d},
\end{equation}
where $\phi(\cdot)$ maps lesion embeddings $\mathbf{v}_\ell$ into the diagnostic
evidence embedding space.

Let $\mathbf{Y} \in \{0,1\}^{N_I \times N_R}$ denote the sparse binary relation
matrix induced by paired image--report samples, where $\mathbf{Y}_{ij}=1$
indicates that $\mathbf{I}_i$ and $\mathbf{R}_j$ form a paired instance.
When paired supervision is available, we can directly set $\mathbf{P}=\mathbf{Y}$,
leading to evidence-aware contrastive learning with hard positives.

\begin{figure}[t]
	\centering
	\includegraphics[scale=.5]{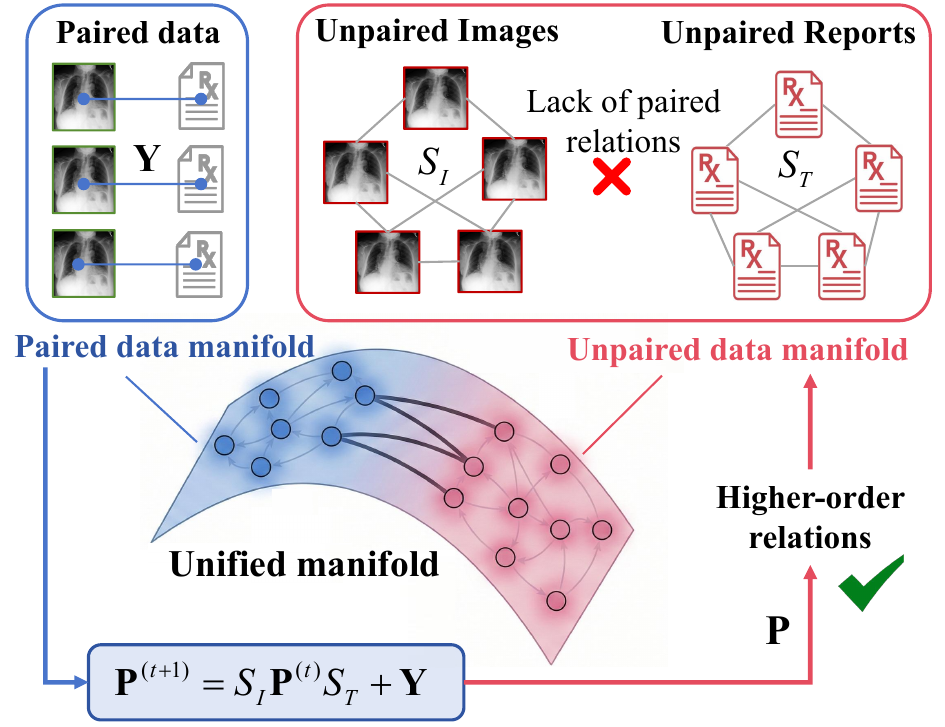}
	\caption{
Evidence-guided higher-order alignment under scarce paired supervision.
Sparse paired links $\mathbf{Y}$ are grounded in a shared diagnostic evidence space and propagated over intra-modal evidence graphs to infer  higher-order relations $\mathbf{P}$.
}
	\label{fig:manifold}
\end{figure}

However, under scarce paired supervision, $\mathbf{Y}$ is extremely sparse and
cannot provide sufficient alignment signals.
To address this limitation, we infer higher-order relations by propagating
the sparse seed relations over intra-modal evidence graphs, which effectively
transfers supervision from paired data to unpaired unimodal data.

We first construct a report--report evidence graph based on report-level
representations:
\begin{equation}
A_{TT}(i,j) = \mathrm{sim}(\mathbf{H}_{\mathbf{R}_i}, \mathbf{H}_{\mathbf{R}_j}),
\end{equation}
and row-normalize it to obtain the text-side propagation matrix $S_T$.
Likewise, we construct an image--image evidence graph:
\begin{equation}
A_{II}(i,j) = \mathrm{sim}(\mathbf{H}_{\mathbf{I}_i}, \mathbf{H}_{\mathbf{I}_j}),
\end{equation}
yielding the image-side propagation matrix $S_I$ after normalization.

Starting from $\mathbf{P}^{(0)}=\mathbf{Y}$, we perform label propagation to infer
higher-order cross-modal relations:
\begin{equation}
\mathbf{P}^{(t+1)}
=
S_I \mathbf{P}^{(t)} S_T
+
\mathbf{Y},
\label{eq:relation_update}
\end{equation}
where the residual term $\mathbf{Y}$ preserves original paired relations and
stabilizes propagation.
After two propagation steps, we obtain higher-order relations $\mathbf{P}$ and apply
row-wise normalization.
The resulting $\mathbf{P}$ enables us to treat evidence-similar image--report
pairs as weak positives, thereby supporting evidence-aware cross-modal
alignment even when explicit pairing is limited as shown in Figure~\ref{fig:manifold}.

The overall training objective integrates diagnostic evidence modeling
and cross-modal alignment:
\begin{equation}
\mathcal{L}
=
\mathcal{L}_{\text{rec}}
+
 \mathcal{L}_{p}^{\text{evid}}
+
\mathcal{L}_{u}^{\text{evid}}
+
 \mathcal{L}_{\text{evi-align}},
\end{equation}
where $\mathcal{L}_{\text{rec}}$ learns a structured diagnostic evidence space from reports,
$\mathcal{L}_{p}^{\text{evid}}$ aligns the limited paired samples within this diagnostic evidence space,
$\mathcal{L}_{u}^{\text{evid}}$ enforces evidence consistency for unpaired images,
and $\mathcal{L}_{\text{evi-align}}$ performs evidence-guided cross-modal alignment.
All algorithms are described as follows.

\begin{algorithm}[!htbp]
\caption{LGDEA: LLM-Guided Diagnostic Evidence Alignment}
\label{alg:lgdea}
\small
\begin{algorithmic}[1]
\REQUIRE Paired data $\mathcal{D}_p$, unpaired images $\mathcal{D}_u^I$, unpaired reports $\mathcal{D}_u^R$.
\REQUIRE Encoders $f_{\text{image}}, f_{\text{text}}$, prototypes $\{\mu_k\}_{k=1}^K$, \\lesion queries $\{q_\ell\}_{\ell=1}^L$.
\ENSURE Trained model parameters.

\WHILE{not converged}
\STATE Sample a mini-batch of paired and \\unpaired images/reports.

\vspace{0.2em}
\STATE \textbf{(1) Build diagnostic evidence space.}
\STATE Extract evidence $\mathcal{E}(R)=\mathrm{LLM}(R)$ and encode $\{z_n\}$.
\STATE Update evidence prototypes by minimizing $\mathcal{L}_{\text{rec}}$.

\vspace{0.2em}
\STATE \textbf{(2) Learn lesion-level visual evidence.}
\STATE Obtain lesion embeddings $\{v_\ell\}$ using lesion queries and attention.
\STATE Project lesions into prototype space to get $\bar{Q}_I$.
\STATE For paired samples, align with report evidence via $\mathcal{L}_{p}^{\text{evid}}$.
\STATE For unpaired images, enforce lesion consistency via $\mathcal{L}_{u}^{\text{evid}}$.

\vspace{0.2em}
\STATE \textbf{(3) Infer higher-order cross-modal relations.}
\STATE Compute image/report evidence representations $\{H_I\},\{H_R\}$.
\STATE Construct evidence graphs $S_I,S_T$ and propagate \\seed links:
$\mathbf{P}^{(t+1)} = S_I \mathbf{P}^{(t)} S_T + \mathbf{Y}$.

\vspace{0.2em}
\STATE \textbf{(4) Evidence-guided weakly-supervised alignment.}
\STATE Optimize $\mathcal{L}_{\text{evi-align}}$ using $\mathbf{P}$-weighted contrastive loss.

\vspace{0.2em}
\STATE Update parameters by minimizing
$\mathcal{L}=\mathcal{L}_{\text{rec}}+\mathcal{L}_{p}^{\text{evid}}+\mathcal{L}_{u}^{\text{evid}}+\mathcal{L}_{\text{evi-align}}$.
\ENDWHILE
\end{algorithmic}
\end{algorithm}

\section{Experiment}
\subsection{Experimental Settings}
\label{sec:exp_setting}

We evaluate LGDEA under both single-domain and cross-domain medical vision--language pretraining settings.
Below we briefly describe the training data, downstream evaluation benchmarks, and baselines.

\paragraph{Training Data}
We conduct pretraining primarily on the MIMIC-CXR dataset~\cite{johnson2019mimic}, which contains substantial chest X-ray images and corresponding radiology reports.
To simulate realistic scenarios with limited pairing, we randomly sample different proportions of paired image--report data, while treating the remaining images and reports as unpaired unimodal data.
For cross-domain pretraining, we additionally incorporate images from CheXpert~\cite{irvin2019chexpert} as unpaired visual data.

\textbf{MIMIC-CXR}
MIMIC-CXR contains 377,110 chest X-ray images and 227,835 radiology reports from 65,379 patients~\cite{johnson2019mimic}.
We use the \textit{Findings} and \textit{Impression} sections as textual input.
Following MedCLIP~\cite{MedCLIP}, the dataset is split into a pretraining set and an evaluation subset (MIMIC-$5{\times}200$).
To simulate limited pairing, we randomly sample 5\% and 10\% of paired image--report data, while treating the remaining images and reports as unpaired unimodal data.
The dataset is publicly available at \url{https://physionet.org/content/mimic-cxr-jpg/2.1.0/}.

\textbf{CheXpert}
CheXpert contains 224,316 chest radiographs from 65,240 patients~\cite{irvin2019chexpert}.
In cross-domain experiments, CheXpert images are used as additional unpaired visual data, combined with paired and unpaired text data from MIMIC-CXR.
The dataset is available at \url{https://www.kaggle.com/datasets/ashery/chexpert}.

\begin{table*}[!htbp]
\renewcommand{\arraystretch}{1.2} 
\caption{The usage of the dataset employed by the model in pre-training and downstream medical tasks.}
\label{tab:dataset}
\centering
\resizebox{\textwidth}{!}{%
\begin{tabular}{l l l c c c c}
\hline
Dataset   & Size/Images & Annotations  & Pre-training & Phrase Grounding & Image-Text Retrieval & Zero-shot Classification \\
\hline
MIMIC-CXR & 377,110     & Paired images and reports & \checkmark &  &  &  \\
CheXpert  & 224,316     & Images                    & \checkmark &  &  &  \\
MS-CXR    & 206         & Bounding boxes, radiology phrase &  & \checkmark &  &  \\
MIMIC-$\rm{5\times200}$ & 1,000       & Radiology text, 5 diseases       &  &  & \checkmark & \checkmark \\
COVID          & 5,162   & COVID-19/normal labels           &  &  &  & \checkmark \\
RSNA Pneumonia & 12,024  & Pneumonia/normal labels          &  &  &  & \checkmark \\
NIH Chest X-rays& 112,120  & 14 disease/normal labels          &  &  &  & \checkmark \\
\hline
\end{tabular}%
}
\end{table*}

\paragraph{Downstream Tasks}
We evaluate learned representations on a diverse set of downstream medical vision--language tasks, including phrase grounding, image--text retrieval, and zero-shot classification.
Specifically, we use MS-CXR for phrase grounding, MIMIC-$5{\times}200$ for retrieval and zero-shot classification, and RSNA Pneumonia, COVID, and NIH Chest X-rays for disease classification.


\textbf{MS-CXR.}
MS-CXR is a phrase grounding benchmark with radiologist-annotated bounding boxes for eight disease categories~\cite{BioViL}.

\textbf{MIMIC-$5{\times}200$.}
A balanced subset constructed from MIMIC-CXR for zero-shot classification and image--text retrieval, covering five disease categories following GLoRIA~\cite{huang2021gloria}.

\textbf{RSNA Pneumonia.}
A binary chest X-ray dataset annotated for pneumonia detection with bounding boxes~\cite{shih2019augmenting}.
We follow MedCLIP~\cite{MedCLIP} to construct class-balanced training and testing splits.

\textbf{COVID and NIH Chest X-rays.}
We evaluate on the COVID-19 dataset~\cite{rahman2021exploring} and the NIH Chest X-rays dataset~\cite{wang2017chestxray} for disease classification using official splits.

\paragraph{Baselines}
We compare LGDEA with representative state-of-the-art medical VLP methods, including MedCLIP~\cite{MedCLIP}, MGCA~\cite{MGCA}, MedKLIP~\cite{MedKLIP}, CLEFT~\cite{CLEFT}, MAVL~\cite{MAVL}, PRIOR~\cite{PRIOR}, CARZero~\cite{CARZero}, MedAligner~\cite{MedAligner}, and AFLoc~\cite{AFLoc}.
All baselines are trained under consistent data settings for fair comparison.

\paragraph{Implementation Details}
LGDEA is trained with a multi-phase optimization schedule.
The initial training phase uses a batch size of 128 and a learning rate of $5\times10^{-5}$ for 2 epochs,
followed by subsequent phases trained for 5 epochs each with a batch size of 64 and a learning rate of $1\times10^{-4}$.
All experiments use AdamW with a weight decay of $1\times10^{-6}$ and are conducted on two NVIDIA A100 GPUs (40GB).
Images are resized to $256{\times}256$ and randomly cropped to $224{\times}224$ during training.

\subsection{Experimental Results}

We evaluate the proposed LGDEA framework under both single-domain and cross-domain medical vision--language pretraining settings with limited paired image--report. In the single-domain setting, all data are drawn from MIMIC-CXR, where only a small proportion (5\% or 10\%) of image--report pairs is retained and the remaining images and reports are included without explicit cross-modal pairing.
In the cross-domain setting, paired samples and reports are from MIMIC-CXR, while images from CheXpert are incorporated as an additional visual source.
Unimodal reports are derived from the remaining MIMIC-CXR reports.
Across different pairing ratios, LGDEA is evaluated on multiple downstream tasks, including phrase grounding, image–text retrieval and zero-shot classification, assessing fine-grained diagnostic localization, cross-modal semantic alignment, and generalization to unseen diseases and domains.

\begin{table*}[tbp]
\renewcommand{\arraystretch}{1.25}
\caption{The performance of the Phrase Grounding task is evaluated on the MS-CXR dataset using Contrast-to-Noise Ratio (CNR) scores across eight disease categories.}
\label{table1}
\centering

\setlength{\tabcolsep}{4pt}
\begin{adjustbox}{width=\textwidth}
\begin{tabular}{>{\centering\arraybackslash}p{1.3cm}|c|cccccccc}
\toprule
\textbf{Group} & \textbf{Model} 
& Atelectasis & Cardiomegaly & Consolidation & Lung Opacity 
& Edema & Pneumonia & Pneumothorax & \makecell{Pleural\\Effusion} \\
\hline

\multirow{11}{*}{\makecell{Paired\\Baseline}}
& MedCLIP    
& 0.5554 & 0.5253 & 0.5767 & 0.5599 & 0.4433 & 0.6175 & 0.4686 & 0.5076 \\
& GLoRIA     
& 0.6088 & 0.5253 & 0.6194 & 0.5376 & 0.5062 & 0.7308 & 0.5896 & 0.8991 \\
& BioViL     
& 0.5376 & 0.5014 & 0.5852 & 0.5266 & 0.4727 & 0.4602 & 0.4882 & 0.5841 \\
& MGCA       
& 0.6338 & 0.5680 & 0.6543 & 0.5586 & 0.5247 & 0.8324 & 0.5240 & 0.8207 \\
& MedKLIP    
& 0.5570 & 0.5282 & 0.5921 & 0.5926 & 0.4905 & 0.6781 & 0.5858 & 0.6398 \\
& PRIOR      
& 0.6452 & 0.5601 & 0.5876 & 0.6624 & 0.5548 & 0.6693 & 0.6564 & 0.8221 \\
& CLEFT      
& 0.6871 & 0.5687 & 0.6492 & 0.5699 & 0.4989 & 0.7259 & 0.5976 & 0.7974 \\
& MAVL       
& 0.5811 & 0.5417 & 0.5486 & 0.6381 & 0.5115 & 0.7056 & 0.5952 & 0.7601 \\
& CARZero    
& 0.6757 & 0.5214 & 0.6889 & 0.5572 & 0.5209 & 0.8438 & 0.6326 & 0.9672 \\
& AFLoc      
& 0.7941 & 0.5928 & 0.7688 & 0.6894 & 0.6271 & 0.7484 & 0.6810 & 0.9866 \\
& MedAligner 
& 0.8312 & 0.6058 & 0.8144 & 0.7163 & 0.7335 & 0.8489 & 0.7249 & 1.0207 \\
\hline

\multirow{2}{*}{\makecell{Single\\Domain}}
& LGDEA$_{5\%\text{pair}}$  
& 0.8000 & 0.9127 & 0.8071 & 0.9032
& 0.8081 & 0.7849 & 0.7902 & 0.9742 \\
& LGDEA$_{10\%\text{pair}}$ 
& \textbf{0.8449} & \textbf{0.9525} & \textbf{0.8357} & \textbf{0.9564}
& \textbf{0.8172} & \textbf{0.9353} & \textbf{0.8526} & \textbf{1.0928} \\
\hline

\multirow{2}{*}{\makecell{Cross\\Domain}}
& LGDEA$_{5\%\text{pair}}$  
& 0.8011 & 0.9111 & 0.7975 & 0.8789
& 0.7868 & 0.8235 & 0.7067 & 0.6667 \\
& LGDEA$_{10\%\text{pair}}$ 
& 0.8306 & 0.9416 & 0.8126 & 0.9003
& 0.7891 & 0.9214 & 0.7952 & 0.7519 \\
\bottomrule

\end{tabular}
\end{adjustbox}
\end{table*}

\begin{figure}[!htbp]
	\centering
		\includegraphics[scale=.5]{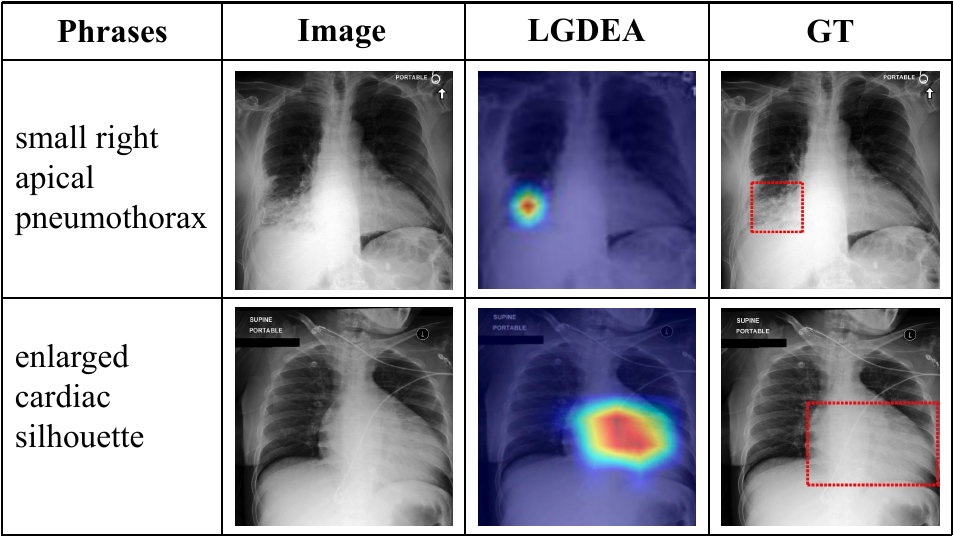}
	\caption{Attention maps for two disease categories are visualized on the MS-CXR dataset, comparing LGDEA with GT. Red bounding boxes indicate the ground truth regions relevant to phrase grounding. Highlighted pixels correspond to higher activation weights, reflecting stronger associations between specific diagnostic terms and image regions.}
	\label{Attention_phrase}
\end{figure}

\begin{figure*}[!htbp]
	\centering
		\includegraphics[scale=.3]{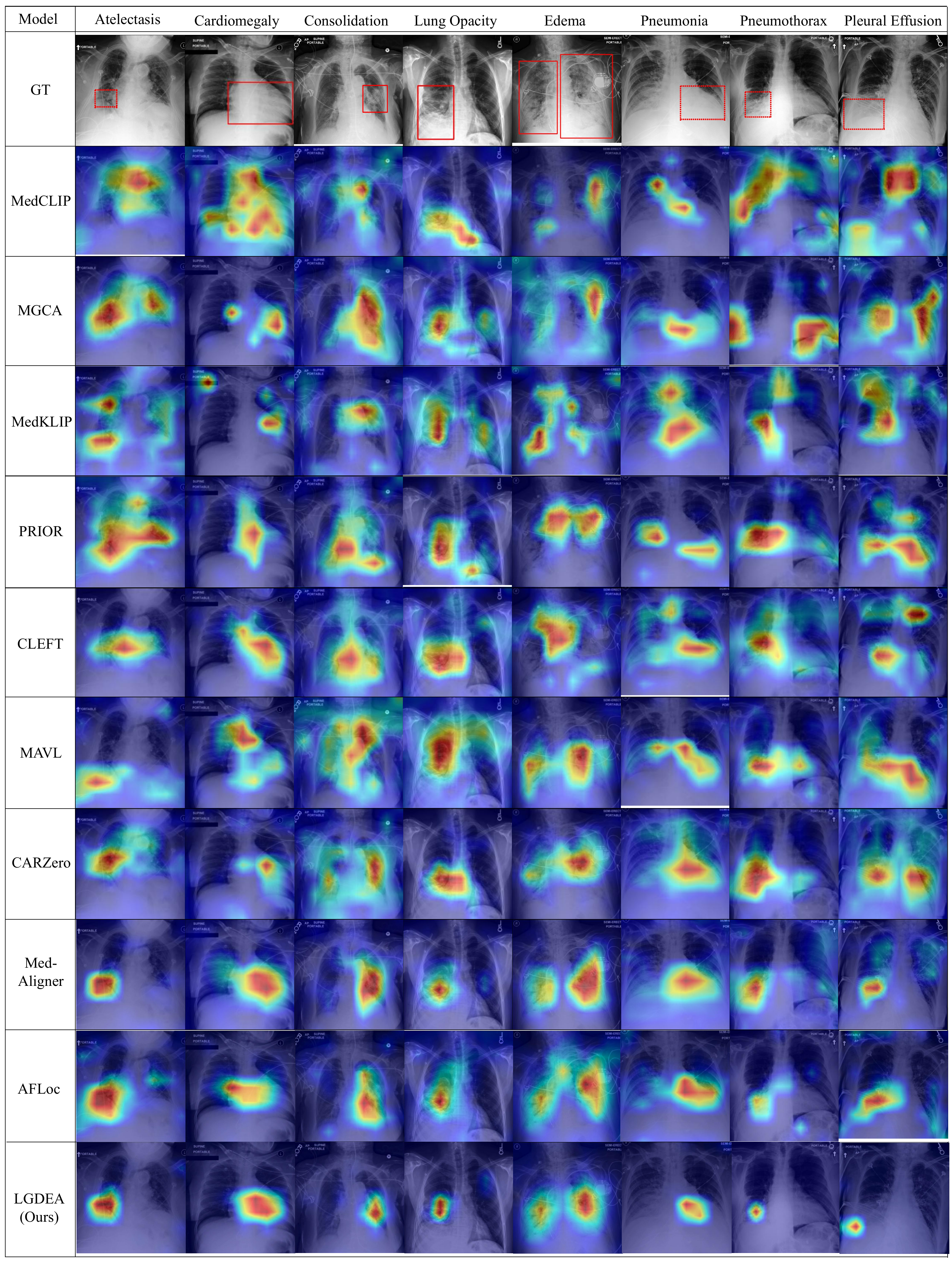}
	\caption{Attention maps for eight disease categories are visualized on the MS-CXR dataset, comparing LGDEA with nine baseline methods. Red bounding boxes indicate the ground truth regions relevant to phrase grounding. Highlighted pixels correspond to higher activation weights, reflecting stronger associations between specific diagnostic terms and image regions.}
	\label{Attention}
\end{figure*}

\textbf{Phrase Grounding}.
Phrase grounding aims to accurately associate textual phrases with their corresponding regions in medical images, thereby enabling fine-grained localization of diagnostic cues and enhancing model interpretability.Table \ref{table1} summarizes the phrase grounding results on the MS-CXR dataset, evaluated using the contrast-to-noise ratio (CNR) \cite{hendrick2008signal}.
With only 10\% of the paired image–text data, LGDEA achieves the highest CNR scores across all eight disease categories, substantially outperforming methods that leverage fully paired image–text relations and focus on local alignment modeling, such as AFLoc and MedAligner.
Even when trained on cross-domain image–text data, LGDEA achieves higher CNR scores across all eight disease categories using only 10\% of the paired image–text samples. 
In addition, visualization results based on Grad-CAM \cite{selvaraju2017grad} heatmaps (Figure \ref{Attention_phrase}) further demonstrate that LGDEA can accurately localize disease-related phrases to the corresponding pathological regions. More visualizations are in the Figure \ref{Attention}.
More importantly, we re-evaluate baseline methods that depend on fully paired supervision under the limited pairing ratios (Table~\ref{pair_methods}), which further demonstrates the superiority of LGDEA in scenarios with scarce paired data.

\begin{table*}[!htbp]
\renewcommand{\arraystretch}{1.25}
\caption{The performance of baselines trained with limited paired image–text data on the phrase grounding task on the MS-CXR dataset, evaluated using the contrast-to-noise ratio (CNR) score across eight disease categories.}
\label{pair_methods}
\centering

\setlength{\tabcolsep}{4pt}
\begin{adjustbox}{width=\textwidth}
\begin{tabular}{>{\centering\arraybackslash}p{1.3cm}|c|cccccccc}
\toprule
\textbf{Group} & \textbf{Model} 
& Atelectasis & Cardiomegaly & Consolidation & Lung Opacity 
& Edema & Pneumonia & Pneumothorax & \makecell{Pleural\\Effusion} \\
\hline

\multirow{5}{*}{\makecell{Single\\Domain}}
& GLoRIA$_{50\%\text{pair}}$   
& 0.4486 & 0.3050 & 0.4359 & 0.4479 & 0.2069 & 0.4941 & 0.3017 & 0.3533 \\
& MGCA$_{50\%\text{pair}}$   
& 0.4620 & 0.2966 & 0.4363 & 0.4897 & 0.2352 & 0.6071 & 0.3569 & 0.4338 \\
& MedAligner$_{50\%\text{pair}}$ 
& 0.6346 & 0.4149 & 0.5344 & 0.5431 & 0.5889 & 0.6836 & 0.6021 & 0.7298 \\

& LGDEA$_{5\%\text{pair}}$  
& 0.8000 & 0.9127 & 0.8071 & 0.9032
& 0.8081 & 0.7849 & 0.7902 & 0.9742 \\
& LGDEA$_{10\%\text{pair}}$ 
& \textbf{0.8449} & \textbf{0.9525} & \textbf{0.8357} & \textbf{0.9564}
& \textbf{0.8172} & \textbf{0.9353} & \textbf{0.8526} & \textbf{1.0928} \\
\hline

\multirow{2}{*}{\makecell{Cross\\Domain}}
& LGDEA$_{5\%\text{pair}}$  
& 0.8011 & 0.9111 & 0.7975 & 0.8789
& 0.7868 & 0.8235 & 0.7067 & 0.6667 \\
& LGDEA$_{10\%\text{pair}}$ 
& 0.8306 & 0.9416 & 0.8126 & 0.9003
& 0.7891 & 0.9214 & 0.7952 & 0.7519 \\
\bottomrule

\end{tabular}
\end{adjustbox}
\end{table*}

\begin{figure}[!htbp]
	\centering
		\includegraphics[scale=.5]{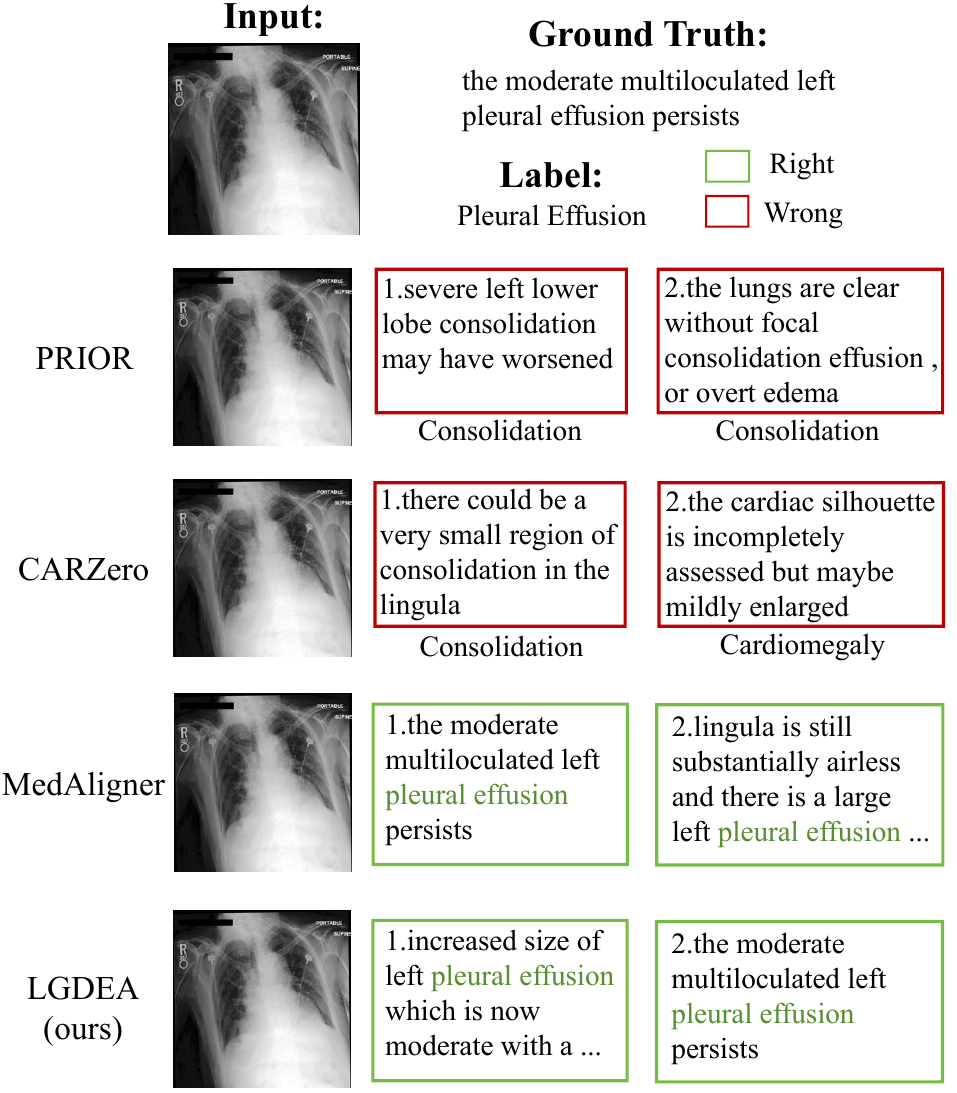}
	\caption{Qualitative results of image-to-text retrieval. We show the  top two retrieved texts for LGDEA, with comparisons to three baselines. We note the categories below wrongly retrieved samples.}
	\label{Attention_retrieval}
\end{figure}

\textbf{Image-Text Retrieval.} To evaluate the matching performance between medical images and textual descriptions, we conducted experiments on the MIMIC-$\rm{5\times200}$ dataset. Specifically, for each query image, we computed the cosine similarity between its [CLS] token representation and the representations of candidate sentences for retrieval. Retrieval performance was assessed using the Precision@K metric.
The results are summarized in Table \ref{Table2}. 
Although PRIOR and CARZero rely on 100\% paired image–text data and benefit from their global image–text alignment to achieve competitive performance, LGDEA, using only 10\% of the paired image–text relations, surpasses them on Precision@1, Precision@2, Precision@5, and Precision@10.
Moreover, with only 10\% of the paired image–text relations, LGDEA outperforms the best-performing MedAligner in terms of Precision@1 and Precision@10, while achieving comparable precision on Precision@2 and Precision@5.
To better illustrate LGDEA semantic understanding capabilities, we present a case study of image-text retrieval examples in Figure \ref{Attention_retrieval}.

\begin{table}[!t]
\centering
\renewcommand{\arraystretch}{1.25}
\caption{The Image--Text Retrieval task is compared with state-of-the-art methods on the MIMIC-\text{5$\times$200} dataset, and Precision (\%) scores are reported.}
\label{Table2}
\footnotesize
\setlength{\tabcolsep}{4pt}

\begin{adjustbox}{scale=1}
\begin{tabular}{c|l|cccc}
\toprule
\textbf{Group} & \textbf{Model} & Prec@1 & Prec@2 & Prec@5 & Prec@10 \\
\hline
\multirow{11}{*}{\makecell{Paired\\Baseline}}
& MedCLIP    & 44.40 & 43.45 & 44.50 & 45.58 \\
& GLoRIA     & 42.38 & 45.65 & 47.74 & 41.98 \\
& BioViL     & 48.81 & 47.88 & 50.05 & 32.88 \\
& MGCA       & 50.06 & 49.77 & 49.05 & 47.53 \\
& MedKLIP    & 50.00 & 50.00 & 49.42 & 50.00 \\
& PRIOR      & 53.19 & 51.72 & 51.89 & 41.69 \\
& CLEFT      & 52.75 & 50.31 & 48.75 & 49.81 \\
& MAVL       & 50.00 & 50.62 & 51.06 & 49.97 \\
& CARZero    & 50.00 & 50.00 & 51.65 & 47.38 \\
& AFLoc      & 54.37 & 52.78 & 49.79 & 43.20 \\
& MedAligner & 55.88 & \textbf{54.08} & \textbf{54.05} & 50.56 \\
\hline
\multirow{2}{*}{\makecell{Single\\Domain}}
& LGDEA$_{5\%\text{pair}}$  & 53.25 & 51.03 & 50.66 & 49.98 \\
& LGDEA$_{10\%\text{pair}}$ & \textbf{56.31} & 53.85 & 53.71 & \textbf{50.65} \\
\hline
\multirow{2}{*}{\makecell{Cross\\Domain}}
& LGDEA$_{5\%\text{pair}}$  & 50.88 & 50.68 & 48.57 & 48.33 \\
& LGDEA$_{10\%\text{pair}}$ & 53.58 & 51.23 & 50.06 & 49.14 \\
\bottomrule
\end{tabular}
\end{adjustbox}
\end{table}

\begin{table}[t]
\centering
\renewcommand{\arraystretch}{1.2} 
\caption{The accuracy (\%) of the Zero-shot Classification is evaluated on the MIMIC-$\rm{5\times200}$, COVID, RSNA Pneumonia, and NIH Chest X-rays datasets.}
\label{Table:Classification}

\begin{adjustbox}{width=\columnwidth}
\begin{tabular}{c|l|cccc}
\toprule
\textbf{Group} & \textbf{Model} & \makecell{MIMIC-\\$\rm{5\times200}$} & COVID & RSNA & \makecell{NIH Chest\\X-rays} \\
\hline

\multirow{11}{*}{\makecell{Paired\\Baseline}}
& MedCLIP    & 52.50 & 77.70 & 79.90 & 58.84 \\
& GLoRIA     & 72.22 & 87.77 & 78.55 & 59.47 \\
& BioViL     & 73.38 & 80.10 & 78.46 & 61.33 \\
& MGCA       & 74.87 & 86.87 & 79.20 & 68.40 \\
& MedKLIP    & 51.94 & 83.26 & 74.65 & 79.40 \\
& PRIOR      & 76.83 & 86.27 & 74.73 & 80.84 \\
& CLEFT      & 75.47 & 84.18 & 78.25 & 78.95 \\
& MAVL       & 74.60 & 83.06 & 78.14 & 82.77 \\
& CARZero    & 76.06 & 86.80 & 78.55 & 83.16 \\
& AFLoc      & 76.20 & 87.80 & 73.52 & 83.87 \\
& MedAligner & 77.89 & 88.20 & \textbf{80.05} & 84.11 \\
\hline
\multirow{2}{*}{\makecell{Single\\Domain}}
& LGDEA$_{5\%\text{pair}}$  & 79.44 & 87.53 & 78.48 & 85.88 \\
& LGDEA$_{10\%\text{pair}}$ & \textbf{80.00} & \textbf{90.47} & 79.26 & \textbf{87.06} \\
\hline
\multirow{2}{*}{\makecell{Cross\\Domain}}
& LGDEA$_{5\%\text{pair}}$  & 74.10 & 88.77 & 77.03 & 77.06 \\
& LGDEA$_{10\%\text{pair}}$ & 76.78 & 90.13 & 78.92 & 78.56 \\
\bottomrule
\end{tabular}
\end{adjustbox}
\end{table}

\begin{table*}[!htbp]
\renewcommand{\arraystretch}{1.3}
\caption{The performance of the phrase grounding task is compared with the all ablated variants, using Contrast-to-Noise Ratio (CNR) scores across eight disease categories.}
\label{tab:ablation_phrase}
\centering
\begin{adjustbox}{scale=0.9}
\resizebox{\textwidth}{!}{%
\begin{tabular}{l | c c c c c c c c}
\toprule
Model & Atelectasis & Cardiomegaly & Consolidation & Lung Opacity & Edema & Pneumonia & Pneumothorax & Pleural Effusion \\
\hline
Full LGDEA                             & 0.8449 & 0.9525 & 0.8357 & 0.9564 & 0.8172 & 0.9353 & 0.8526 & 1.0928 \\
w/o $\mathcal{L}_{\text{rec}}$         & 0.6516 & 0.7170 & 0.6172 & 0.7604 & 0.7236 & 0.8142 & 0.6407 & 0.8125 \\
w/o $\mathcal{L}_{p}^{\text{evid}}$    & 0.7214 & 0.8121 & 0.7210 & 0.8946 & 0.7571 & 0.8856 & 0.7179 & 0.9072 \\
w/o $\mathcal{L}_{u}^{\text{evid}}$    & 0.7064 & 0.7977 & 0.6972 & 0.8828 & 0.7724 & 0.8153 & 0.7348 & 0.8524 \\
w/o $\mathcal{L}_{\text{evi-align}}$   & 0.7355 & 0.8302 & 0.8049 & 0.9150 & 0.7751 & 0.8973 & 0.6238 & 0.9196 \\
\bottomrule
\end{tabular}}
\end{adjustbox}
\end{table*}

\textbf{Zero-Shot Classification.} To assess the adaptability and generalization capability of the learned image encoder, we further evaluated its performance on zero-shot classification tasks. 
On the MIMIC-$\rm{5\times200}$ dataset, we perform classification evaluation by computing the similarity between the labels predicted by the image encoder and the ground-truth labels. For the COVID, RSNA Pneumonia, and NIH ChestX-ray datasets, we attach a classification head to the image encoder and train it using the cross-entropy loss, enabling the model to make predictions solely based on image features. Table \ref{Table:Classification} summarizes the classification results on MIMIC-$\rm{5\times200}$, COVID, RSNA Pneumonia, and NIH ChestX-ray. 
When trained with only 10\% paired image–text data, LGDEA outperforms all baselines across most datasets, demonstrating strong cross-dataset generalization. While it performs slightly worse than MedAligner and MedCLIP on RSNA, it surpasses them on the remaining datasets.

\subsection{Ablation Study}
\label{ablation_com}
\textbf{The contributions of different components.}
We conduct ablation studies to assess the contribution of each core component in LGDEA. All variants are evaluated on the same downstream benchmarks for fair comparison. Specifically, we consider: (1) w/o $\mathcal{L}_{\text{rec}}$, removing diagnostic evidence prototype learning and using global text embeddings instead; (2) w/o $\mathcal{L}_{p}^{\text{evid}}$, removing paired evidence distillation; (3) w/o $\mathcal{L}_{u}^{\text{evid}}$, removing unpaired evidence consistency; and (4) w/o $\mathcal{L}_{\text{evi-align}}$, disabling evidence relation inference and using only paired samples as positives. As shown in Table~\ref{tab:ablation_phrase} and \ref{tab:ablation_retrieval}, removing any component degrades performance, with w/o $\mathcal{L}_{\text{rec}}$ causing the largest drop, highlighting the importance of diagnostic evidence prototype learning under limited paired supervision.

\textbf{The effects of different LLMs.}
In our main experiments, diagnostic evidence is extracted using Spark-Desk. To evaluate the robustness of LGDEA to the choice of evidence extractor, we replace Spark with two open-source LLMs, Qwen-7B \cite{bai2023qwen} and LLaMA-8B \cite{grattafiori2024llama}, while keeping all other components unchanged. As shown in Table ~\ref{Table:LLM_classification} and Table ~\ref{table:LLM_phrase}, LGDEA achieves stable performance across different LLM settings on both phrase grounding and image classification tasks. Although the extracted diagnostic evidence varies in linguistic style across LLMs, the overall performance remains largely consistent, indicating that LGDEA does not rely on a specific LLM but instead benefits primarily from the semantic effectiveness of the diagnostic evidence.

\begin{table}[!htbp]
\renewcommand{\arraystretch}{1.3}
\caption{The Image--Text Retrieval task is compared with the all ablated variants on the MIMIC-$\rm{5\times200}$ dataset, and Precision (\%) scores are reported.}
\label{tab:ablation_retrieval}
\centering
\begin{adjustbox}{scale=1}
\begin{tabular}{l | c c c c}
\toprule
Model & Prec@1 & Prec@2 & Prec@5 & Prec@10  \\
\hline
Full LGDEA                             & 56.31  & 53.85  & 53.71  & 50.65  \\
w/o $\mathcal{L}_{\text{rec}}$         & 51.56  & 50.42  & 48.12  & 45.29  \\
w/o $\mathcal{L}_{p}^{\text{evid}}$    & 54.06  & 52.12  & 49.89  & 46.37  \\
w/o $\mathcal{L}_{u}^{\text{evid}}$    & 53.12  & 51.34  & 50.24  & 47.68  \\
w/o $\mathcal{L}_{\text{evi-align}}$   & 53.25  & 51.68  & 50.47  & 48.61  \\
\bottomrule
\end{tabular}
\end{adjustbox}
\end{table}

\begin{table*}[!htbp]
\renewcommand{\arraystretch}{1.3}
\caption{The performance of the phrase grounding task is compared with LLaMA and Qwen on the MS-CXR dataset, using Contrast-to-Noise Ratio (CNR) scores across eight disease categories.}
\label{table:LLM_phrase}
\centering
\begin{adjustbox}{scale=1}
\resizebox{\textwidth}{!}{%
\begin{tabular}{l | c c c c c c c c}
\toprule
Model & Atelectasis & Cardiomegaly & Consolidation & Lung Opacity & Edema & Pneumonia & Pneumothorax & Pleural Effusion \\
\hline
LLaMA  & 0.8000 & 0.9443 & 0.8343 & 0.9235 & 0.8115 & 0.9271 & 0.6894 & 0.9328 \\
Qwen   & 0.8182 & 0.9469 & 0.8314 & 0.9160 & 0.8020 & 0.8805 & 0.8521 & 0.8003 \\
\hline
Ours & \textbf{0.8449} & \textbf{0.9525} & \textbf{0.8357} & \textbf{0.9564} & \textbf{0.8172} & \textbf{0.9353} & \textbf{0.8526} & \textbf{1.0928} \\
\bottomrule
\end{tabular}}
\end{adjustbox}
\end{table*}

\begin{table}[!t]
\centering
\renewcommand{\arraystretch}{1.2} 
\caption{The accuracy (\%) of the Zero-shot Classification is evaluated on the MIMIC-$\rm{5\times200}$, COVID, RSNA Pneumonia, and NIH Chest X-rays datasets and compared with LLaMA and Qwen.}
\label{Table:LLM_classification}
\begin{adjustbox}{scale=1, center}
\begin{tabular}{l|cccc}
\toprule
\textbf{Model}    & \makecell{MIMIC-\\$\rm{5\times200}$}  & COVID  & RSNA  &\makecell{NIH Chest\\X-rays}   \\
\hline
LLaMA   & 79.72 & 90.10 & \textbf{79.60} & 85.80  \\
Qwen    & 79.92 & 90.39 & 79.15 & 86.14  \\
Ours    & \textbf{80.00} & \textbf{90.47} & 79.26 & \textbf{87.06}  \\
\bottomrule
\end{tabular}
\end{adjustbox}
\end{table}

\subsection{Hyper-parameter Analysis}
We analyze the sensitivity of LGDEA to two key hyper-parameters: the number of diagnostic prototypes $K$, which controls the semantic capacity of the shared evidence space, and the number of lesion queries $L$, which determines the granularity of lesion-level visual modeling.

\begin{figure}[h]
\vskip 0.2in
\begin{center}
\centerline{\includegraphics[width=\columnwidth]{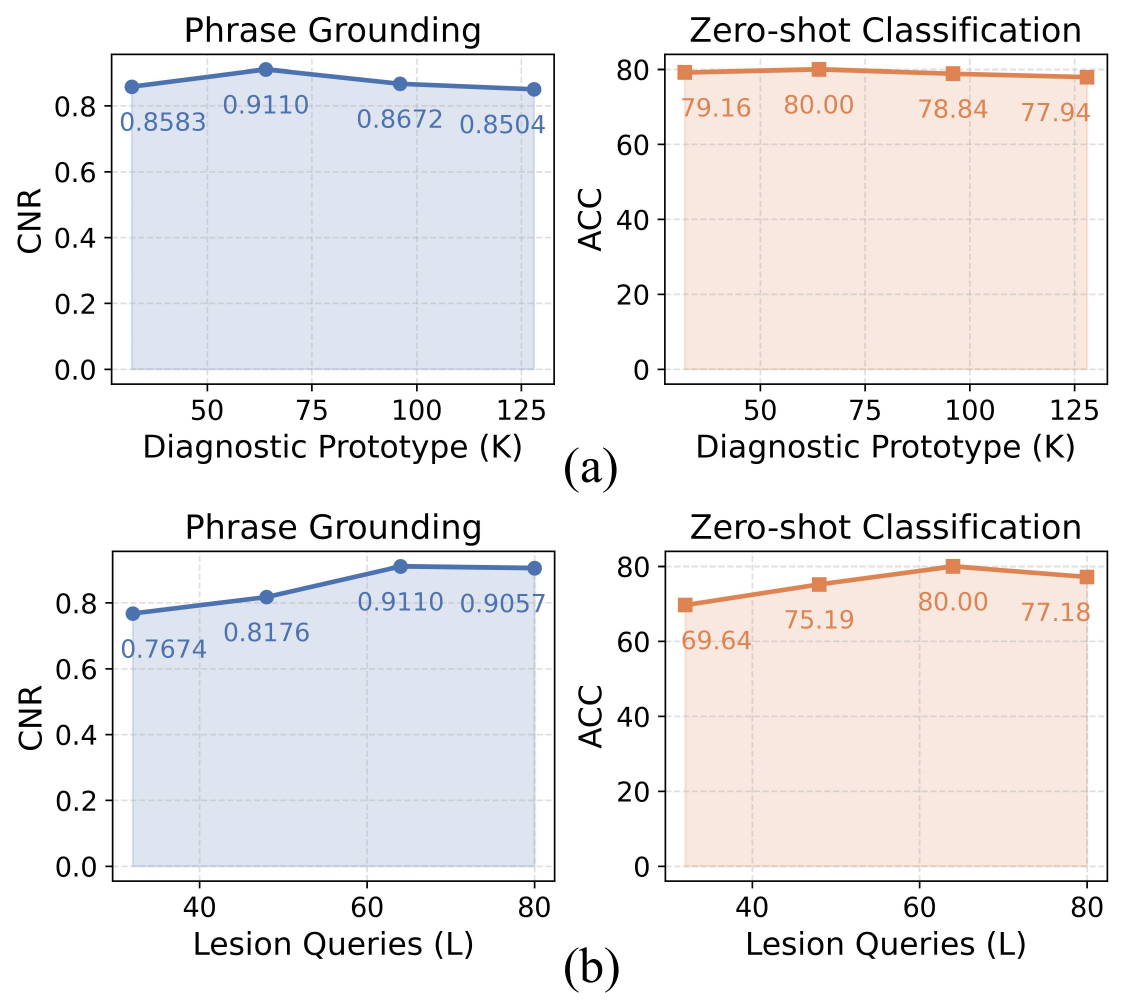}}
\caption{
(a) Effect of the number of diagnostic prototypes $K$ and
(b) effect of the number of lesion queries $L$,
evaluated using average CNR on the MS-CXR phrase grounding task
and ACC on the MIMIC-$5\times200$ zero-shot classification task.
}
\label{fig:iter}
\end{center}
\vskip -0.2in
\end{figure}

\textbf{Number of diagnostic prototypes $K$.}
We vary $K \in \{32, 64, 96, 128\}$ while fixing $L=64$. 
As shown in Figure~\ref{fig:iter} and Table~\ref{prototype}, 
a small $K$ limits semantic expressiveness by forcing heterogeneous evidence to share prototypes, degrading phrase grounding and zero-shot classification performance. Increasing $K$ improves fine-grained evidence–lesion alignment, while overly large values lead to performance saturation and slight degradation. Overall, $K=64$ provides the best trade-off between expressiveness and stability and is used in all experiments.

\textbf{Number of lesion queries $L$.}
We fix $K=64$ and vary $L \in \{32, 48, 64, 80\}$. 
As shown in Figure~\ref{fig:iter} and Table~\ref{query}, 
increasing $L$ enhances coverage of multiple pathological regions and improves grounding accuracy. However, excessively large $L$ introduces redundant and overlapping queries, leading to slightly degraded performance. We therefore adopt $L=64$ as a balanced setting across all downstream tasks.

\begin{table}[!htbp]
\renewcommand{\arraystretch}{1.2}
\caption{Image classification accuracy (\%) on the COVID, RSNA, and NIH Chest X-rays datasets with varying numbers of diagnostic prototypes $K$, evaluating the effect of semantic capacity in the shared evidence space.}
\label{prototype}
\centering
\begin{adjustbox}{width=\linewidth}
\begin{tabular}{
>{\centering\arraybackslash}p{0.42\linewidth} |
>{\centering\arraybackslash}p{0.16\linewidth}
>{\centering\arraybackslash}p{0.16\linewidth}
>{\centering\arraybackslash}p{0.22\linewidth}
}
\toprule
Number of diagnostic prototypes $K$ & COVID & RSNA & NIH Chest X-rays \\
\midrule
32   & 87.14          & 87.14          & 85.91  \\
64   & \textbf{90.47} & 79.26          & \textbf{87.06}  \\
96   & 86.80          & \textbf{79.37} & 86.88  \\
128  & 84.72          & 78.21          & 86.59  \\
\bottomrule
\end{tabular}
\end{adjustbox}
\end{table}

\begin{table}[!htbp]
\renewcommand{\arraystretch}{1.2}
\caption{Image classification accuracy (\%) on the COVID, RSNA, and NIH Chest X-rays datasets with different numbers of lesion queries $L$, analyzing the impact of lesion-level visual modeling granularity.}
\label{query}
\centering
\begin{adjustbox}{width=\linewidth}
\begin{tabular}{
>{\centering\arraybackslash}p{0.42\linewidth} |
>{\centering\arraybackslash}p{0.16\linewidth}
>{\centering\arraybackslash}p{0.16\linewidth}
>{\centering\arraybackslash}p{0.22\linewidth}
}
\toprule
Number of lesion queries $L$ & COVID & RSNA & NIH Chest X-rays \\
\midrule
32   & 85.10          & 77.71          & 85.06  \\
48   & 86.73          & 78.95          & 85.53  \\
64   & \textbf{90.47} & \textbf{79.26} & \textbf{87.06}  \\
80   & 85.57          & 79.16          & 86.72  \\
\bottomrule
\end{tabular}
\end{adjustbox}
\end{table}

\section{Conclusion}
In this paper, we show that CLIP-style global and local alignment becomes unreliable for medical vision--language pretraining under scarce paired data. 
To address this issue, we propose LGDEA, an LLM-guided framework that shifts pretraining from feature-level alignment to diagnostic evidence alignment. 
LGDEA extracts key evidence from radiology reports and guides lesion-level visual evidence learning to align images with clinically meaningful semantics. 
Extensive experiments on phrase grounding, image--text retrieval, and zero-shot classification demonstrate consistent gains, and LGDEA can rival methods pretrained with substantially more paired data.

\bibliographystyle{IEEEtran}
\bibliography{references}

\end{document}